\documentclass[10pt,twocolumn,letterpaper]{article}

\usepackage{cvpr}
\usepackage{times}
\usepackage{epsfig}
\usepackage{graphicx}
\usepackage{amsmath}
\usepackage{arydshln}
\usepackage{xcolor}
\usepackage{amssymb}
\usepackage{caption}
\usepackage[pagebackref=true,breaklinks=true,letterpaper=true,colorlinks,bookmarks=false]{hyperref}

\cvprfinalcopy 

\def\cvprPaperID{1089} 

\ifcvprfinal\fi
\begin{document}

\title{Interpreting Super-Resolution Networks with Local Attribution Maps}

\author{Jinjin Gu\\
School of Electrical and Information Engineering,\\
The University of Sydney.\\
{\tt\small jinjin.gu@sydney.edu.au}
\and
Chao Dong\\
Shenzhen Institutes of Advanced Technology,\\
Chinese Academy of Sciences.\\
{\tt\small chao.dong@siat.ac.cn}
}

\maketitle

\begin{abstract}
Image super-resolution (SR) techniques have been developing rapidly, benefiting from the invention of deep networks and its successive breakthroughs.
However, it is acknowledged that deep learning and deep neural networks are difficult to interpret.
SR networks inherit this mysterious nature and little works make attempt to understand them.
In this paper, we perform attribution analysis of SR networks, which aims at finding the input pixels that strongly influence the SR results.
We propose a novel attribution approach called local attribution map (LAM), which inherits the integral gradient method yet with two unique features.
One is to use the blurred image as the baseline input, and the other is to adopt the progressive blurring function as the path function. 
Based on LAM, we show that:
(1) SR networks with a wider range of involved input pixels could achieve better performance. 
(2) Attention networks and non-local networks extract features from a wider range of input pixels.
(3) Comparing with the range that actually contributes, the receptive field is large enough for most deep networks.
(4) For SR networks, textures with regular stripes or grids are more likely to be noticed, while complex semantics are difficult to utilize.
Our work opens new directions for designing SR networks and interpreting low-level vision deep models.

\end{abstract}

\section{Introduction}
Deep learning has recently shown an explosive popularity in the field of image super-resolution (SR) due to its superior performance and flexibility.
\let\thefootnote\relax\footnotetext{Code, more visualization results, and an online demo are available at the project page \url{https://x-lowlevel-vision.github.io/lam.html}.}
Various SR networks have been proposed to learn effective and abstract representations for SR, expecting to continuously improve the SR performance.
Despite their success, these SR networks remain mysterious because what has been learned and how it contributes to their performance remain unclear.
For example, whether larger receptive fields and multi-scale structures are effective for SR networks?
Why the attention and non-local schemes can help improve the SR results?
How could different network architectures affect the information usage and the final performance?
We lack a systematic understanding, and even research tools, towards these open questions.

\begin{figure}[t]
    \centering
    \includegraphics[width=1.0\linewidth]{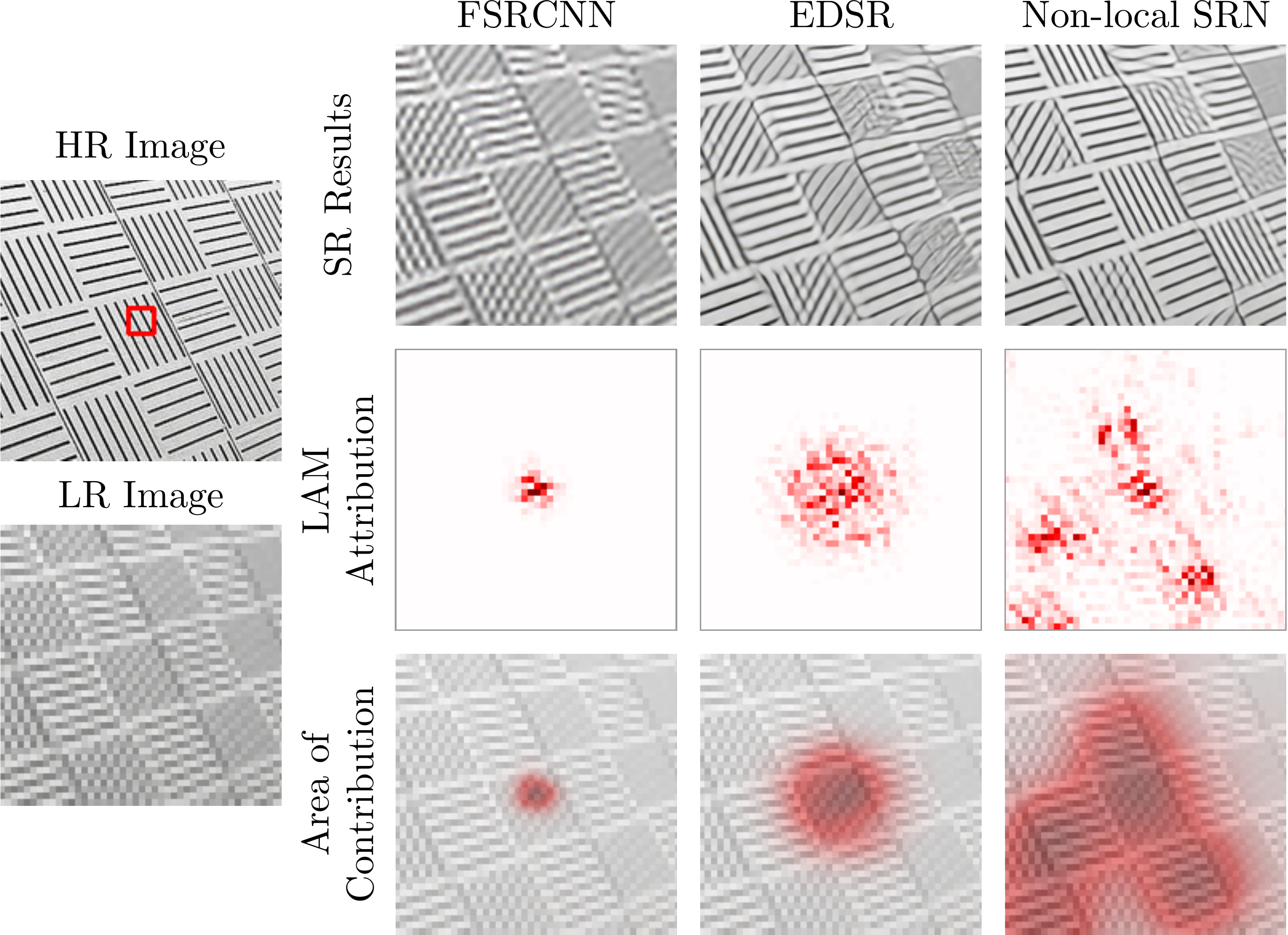}
    \caption{Demo results of the proposed attribution method LAM for SR network interpretation. The LAM maps represent the importance of each pixel in the input LR image \wrt the SR of the patch marked with a red box. We also illustrate the area of contribution in the third line. Our LAM results indicate that FSRCNN \cite{fsrcnn2016} only utilizes very limited information, EDSR \cite{edsr2017} increases the range of information utilization but still fails to reconstruct accurate texture, and non-local network RNAN \cite{nonResidual} can utilize a wider range of information for better SR result.}
    \label{fig:intro}
    \vspace{-6mm}
\end{figure}

In this paper, we propose to conduct attribution analysis of SR networks, aiming at finding input features that strongly influence the network outputs.
The results are often visualized in the attribution maps where the most important pixels are highlighted.
In this manner, we can analyze the pattern of information usage for SR networks, and evaluate whether a SR network could extract high-level semantic information. 
We show some representative results in \figurename~\ref{fig:intro}.
In contrast to the attribution methods that are widely studied in classification networks, the SR networks have not witnessed the development and application of attribution techniques.
As the first work that tries to build attribution method for SR networks, we need to introduce several auxiliary principles.
First, we argue that the attribution should be conducted in local patches rather than global images.
Second, we should analyze patches that are difficult to reconstruct.
Third, we propose to interpret the existence of specific features instead of pixel intensities.
With these preliminaries, we present a new Local Attribution Map (LAM) to interpret SR networks.
LAM employs path integral gradients to conduct attribution analysis.
We use the blurred image as the baseline input and propose a novel progressively blurring function as the path function.
These two strategies are specially designed for SR networks.

Using the proposed LAM method, we draw the following observations:
\textbf{(1)} The range of pixels involved in the SR process has significant impact to the final SR performance. SR networks with a wider range of involved pixels could achieve better performance. Thus deepening or widening the SR network could lead to better performance (under sufficient training).
\textbf{(2)} Attention and non-local schemes could help SR networks extract features from a wider range of input pixels.
\textbf{(3)} For most deep networks, the receptive field is much larger than the pixel range that actually contributes to the SR. Simply expanding the receptive field will not involve more influential pixels.
\textbf{(4)} For existing SR networks, textures with regular stripes or grids are more likely to be extracted, while complex semantics are difficult to extract and utilize.

The above observations are beneficial to SR research both scientifically and practically.
The reasons are at least threefold, with the value of diagnostic for SR networks being the first.
Understanding the mechanism that underlies these SR networks would intuitively tell us the effectiveness of the proposed method.
Second, it would also have direct consequences on designing more efficient architectures.
Third, studying the interpretability of SR networks would also advance the science of deep learning applied to low-level vision tasks.
In light of the above discussion, we hope our work can serve as a stepping stone towards a more formal understanding of SR networks.

\section{Related Work}
\noindent\textbf{SR networks.}\quad
%
%
Since Dong \etal \cite{srcnn2014} introduce the first SR network, plenty of deep learning based SR methods have been proposed, including deeper networks \cite{fsrcnn2016,vdsr2016,espcn2016}, residual architectures \cite{srgan2017,wang2018esrgan}, recurrent architectures \cite{drcn2016,drrn2017}, and attention mechanism \cite{rcan2018,san2019}.
Recently, non-local operations \cite{wang2018non} are also introduced to SR networks \cite{nonRecurrent,nonResidual,crossNonlocal}, expecting to capture long-range information in the image and utilize self-similarity prior of natural images.
The whirlwind of progress in deep learning has delivered remarkable performance in SR field.
Research on SR networks focuses on measuring the difference in performance, lacking a formal understanding of their underlying mechanisms.

\noindent\textbf{Network interpretation.}\quad
Our work is also closely related to network interpretation.
Since the deep networks were applied to computer vision, network interpretation follows a long line of works on understanding the predictions given by the models.
Some works attempt to explore inside the deep network and visualize the learned knowledge, such as natural pre-image \cite{mahendran2016visualizing} and network dissection \cite{zhou2018interpreting}.
%
%
Another straightforward idea, which is more related to our work, is to visualize what part of the input is responsible for the model’s prediction.
These visualization results are called attribution maps. 
%
%
In recent years, various attribution methods have been proposed \cite{simonyan2013deep,springenberg2015striving,sundararajan2017axiomatic,shrikumar2017learning,lundberg2017unified}, aiming at obtaining human-understandable representations for attribution.
Some methods employ networks’ interior activations to localize the discriminative image regions \cite{zhou2016learning,selvaraju2017grad}.
While the above works require the mathematical details of the model, there are works that treat deep models as black-boxes.
These methods usually localize the discriminative image regions by performing perturbation to the input \cite{fong2017interpretable,fong2019understanding}.
%
%
%
In addition, there are also works on interpreting generation networks, \eg, Bau \etal \cite{bau2018gan} propose a framework for visualizing and understanding the structure learned by a generative network, and Shen \etal \cite{shen2020interpreting} describe that a well-trained generative network learns a disentangled representation.

\noindent\textbf{Difference from the previous interpretation research.}\quad
Despite the above works on explaining both discriminative and generative networks, to the best of our knowledge, the SR networks have not witnessed the effort of explanation and visualization.
Interpreting SR networks is different from previous works in both perspectives and approaches.
First, the concerns of security and human-interactable are the key motivations behind the research of classification network interpretation.
However, the motivation of interpreting SR networks lies in pursuing better understanding of the existing models' success and obtaining insights to break performance bottlenecks.
Second, the characteristics of the SR network make the existing interpretation research invalid for SR networks.
New principles and approaches for interpretation need to be established, which is the primal goal of this paper.

\section{Method}
Before diving into the specific method, we first introduce several important auxiliary principles for SR network interpretation in Sec~\ref{sec:mtd:formulations}.
We then briefly describe some prior works of attribution methods in Sec~\ref{sec:mtd:related}.
In Sec~\ref{sec:mtd:lam}, we propose the Local Attribution Map to interpret SR networks.
At last, we provide supplementary discussions in Sec~\ref{sec:mtd:disc}.

\subsection{Auxiliary Principles of Interpretation}
\label{sec:mtd:formulations}
\noindent\textbf{Interpreting local not global.}\quad
Different from classification networks that output predicted labels \wrt the whole image, SR networks output the SR image that are spatially corresponding to the input image.
The independence of SR in different locations poses challenges for interpreting SR networks globally.
We propose to interpret the SR networks in a local manor for specific locations and surroundings. This is consistent with the commonly-used qualitative evaluation strategy, which focuses on the reconstruction of edges and textures. 
We aim to find the input pixels that effectively contribute to the reconstruction of a certain location/area in the output image.
This process can tell us what information has been utilized by the SR network.

\noindent\textbf{Interpreting hard not simple.}\quad
The flat areas and simple edges are relatively easy to reconstruct in SR.
The interpretation of these areas is of limited help in understanding SR networks.
What we really care about is the areas that limit the development of SR.
In these areas, the low-resolution (LR) image usually contains limited information, and different SR networks have distinguishable performance.
Understanding how these SR networks utilize information to obtain different performances could provide useful insights for designing better algorithms.

\noindent\textbf{Interpreting features not pixels.}\quad
When conducting attribution analysis of classification networks, the gradients are usually calculated directly on the predicted probabilities of labels.
However, the outputs of SR networks are pixel intensities, which are strongly correlated with the pixel intensities of the corresponding location in the LR image.
The direct attribution results will also be correlated with the pixel intensities, as they provide the main gradients.
But the intensities, ranging from 0 to 255, could provide little help for network interpretation. 
In contrast, we propose to detect the existence of specific local features, such as edges and textures.
We convert the problem of attribution into \textit{whether there exists edges/textures or not}, instead of why these pixels have such intensities.
In this manner, the attribution results are robust to the brightness changes.

\subsection{Investigating Attribution Methods}
\label{sec:mtd:related}
Before introducing our method, we briefly summarize the recent progress of attribution methods in classification networks. 
Consider an input image $I\in\mathbb{R}^d$ and a classification model $S:\mathbb{R}^d\mapsto\mathbb{R}$, an attribution method provides attribution maps $\mathsf{Attr}_S:\mathbb{R}^d\mapsto\mathbb{R}^d$ for $S$ that are of the same size as the inputs and present the importance for each dimension.
As the most intuitive idea, the gradient for an input $\mathsf{Grad}_S(I)=\frac{\partial S(I)}{\partial I}$ quantifies how much a change in each input dimension would change the output in a small neighborhood around the input \cite{simonyan2013deep,baehrens2010explain}.
However, the gradient method suffers from the ``saturation'' issue \cite{sundararajan2017axiomatic,sturmfels2020visualizing}, which could lead to the problem -- the gradients have small magnitudes and fail to indicate important features.
The element-wise product of the input and the gradient $I\odot\frac{\partial S(I)}{\partial I}$ is proposed to address the saturation problem and reduce visual diffusion \cite{shrikumar2016not}.
Sundararajan \etal \cite{sundararajan2017axiomatic} also propose Integrated Gradients (IG) to alleviate gradient saturation, which is formulated as $\mathsf{IG}_S(I)=(I-I')\cdot\int_0^1\frac{\partial S(I'+\alpha(I-I'))}{\partial I}d\alpha$, where $I'$ is the baseline input that represents the absence of important features.
The idea of introducing baseline inputs is also revealed in many prior works \cite{shrikumar2016not,binder2016layer}.
When we assign blame to a certain cause, we implicitly consider the absence of the cause as a baseline for comparison.
In addition to the above attribution methods, Guided Backpropagation \cite{springenberg2015striving} and Smooth Gradient \cite{smilkov2017smoothgrad} are also widely used in visualizing classification networks.
More information about these methods and their relationship to our work can be found in Appendix~\ref{sec:apd:overview}.

\begin{figure}[t]
    \centering
    \includegraphics[width=\linewidth]{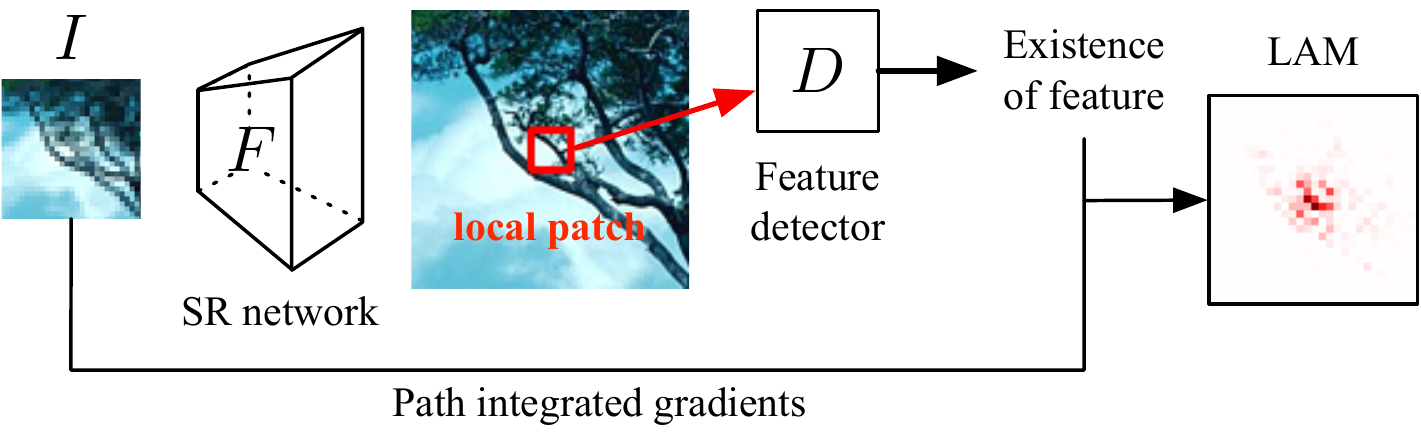}
    \vspace{-8mm}
    \caption{Illustration of the Local Attribution Map (LAM).}
    \label{fig:framework}
    \vspace{-4mm}
\end{figure}  

\begin{figure*}
    \centering
    \includegraphics[width=\linewidth]{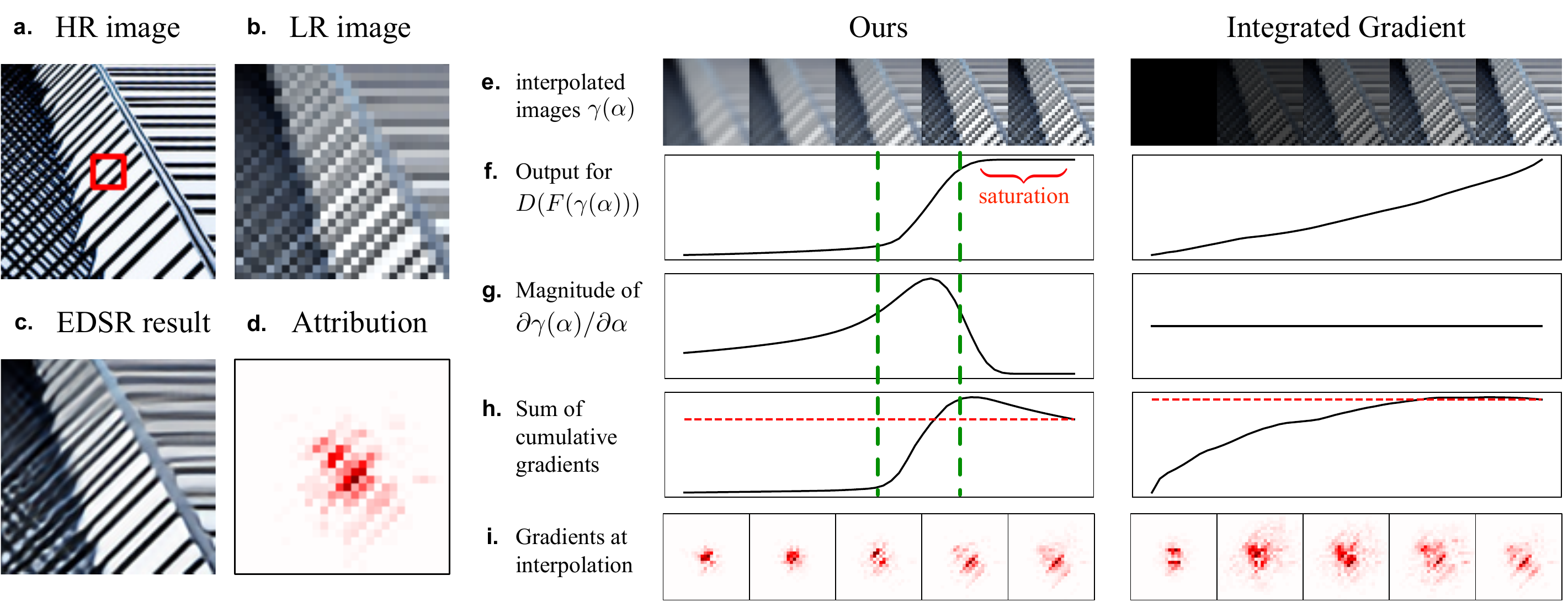}
    \vspace{-6mm}
    \caption{Comparison of different baseline images and path functions. \textbf{(a)} -- \textbf{(d)} are the HR image with the selected local patch highlighted, the LR image, the SR result using EDSR \cite{edsr2017}, and the attribution map using the proposed LAM method, respectively.
    On the right of the figure:
    \textbf{(e)} The images interpolated from the baseline image to the LR image using different baselines and path functions.
    \textbf{(f)} The curves of $D(F(\gamma(\alpha)))$ \wrt $\alpha$. The saturation triggered when the curve shows a flat trend.
    \textbf{(g)} The curves illustrate magnitudes of $\partial\gamma(\alpha)/\partial\alpha$ with different $\alpha$s. Notice high magnitude gradients accumulate at position that $D(F(\gamma(\alpha)))$ changes sharply.
    \textbf{(h)} shows the curve by accumulating gradients over $\alpha$, the red line refers to $D(F(I)) - D(F(I'))$.
    \textbf{(i)} shows the gradients at interpolations.}
    \label{fig:bsl&path}
    \vspace{-4mm}
\end{figure*}  

\subsection{Local Attribution Maps}
\label{sec:mtd:lam}
In this section, we describe the proposed Local Attribution Maps (LAM), which is based on the integrated gradients method \cite{sundararajan2017axiomatic}.
Let $F: \mathbb{R}^{h\times w}\mapsto\mathbb{R}^{sh\times sw}$ be an SR network with the upscaling factor $s$.
As stated before, we interpret $F$ by attributing the existence of certain features of local patches in the output image, instead of the pixel intensities.
We quantify the existence of a specific feature in an $l\times l$ patch located in $(x, y)$ with a detector $D_{xy}:\mathbb{R}^{l\times l}\mapsto\mathbb{R}$.
Here, $D_{xy}$ is implemented by simple operators or filters, which can be easily understood, to avoid introducing additional difficulty to the interpretation. 
In this work, we mainly use the gradient detector to quantify the existence of edges and textures, as 
\begin{equation}
    D_{xy}(I)=\sum_{i\in[x,x+l],j\in[y,y+l]} \nabla_{ij} I,
\end{equation}
where the subscript $i,j$ indicates the location coordinates.
In the following text, we omit the subscripts for convenience without loss of generality.
Given $I\in\mathbb{R}^{h\times w}$ as the input LR image, conducting attribution analysis of an SR network also requires a baseline input image $I'$, which satisfies that $F(I')$ absent certain features existed in $F(I)$.
A simple example of $I'$ is a black image with all zero-value pixels.
Accordingly, $D(F(I))$ will show large numerical advantage compared to $D(F(I'))$.

To obtain the attribution map for $D(F(I))$, we calculate the path integrated gradient along the gradually changing path from $I'$ to $I$.
We represent the path by a smooth path function $\gamma(\alpha):[0,1]\mapsto\mathbb{R}^{h\times w}$, where $\gamma(0)=I'$ and $\gamma(1)=I$.
Then, the $i$th dimension of the local attribution map is defined as follows:
\begin{equation}
    \mathsf{LAM}_{F,D}(\gamma)_i:=\int_{0}^{1}\frac{\partial D(F(\gamma(\alpha))}{\partial\gamma(\alpha)_i}\times\frac{\partial\gamma(\alpha)_i}{\partial\alpha}d\alpha.
    \label{eq:PathIntegratedGradient}
\end{equation}
Obviously, for different baselines and path functions, we obtain different attribution maps.
When the baseline inputs are black images and the path function is linear interpolation, Eq~\eqref{eq:PathIntegratedGradient} degrades to the standard formulation of IG \cite{sundararajan2017axiomatic}.
However, as revealed in \cite{sturmfels2020visualizing}, the choice of baseline and path function may greatly affect the final attribution results.
In this work, we carefully design the baseline inputs and the path function specially for SR networks.

As stated above, a baseline input is meant to represent the ``absence'' of some input features.
In SR, the low-frequency components of the LR image (e.g., color and brightness) contribute less to the final SR performance.
In contrast, the high-frequency components (e.g., edges and textures) are of great importance to achieve good SR results. 
In this work, we design the baseline input by eliminating the high-frequency components.
In implementation, we set it as the blurred version of the LR image, denoted as $I'=\omega(\sigma)\otimes I$, where $\otimes$ represents convolution operation and $\omega(\sigma)$ is the Gaussian blur kernel parameterized by the kernel width $\sigma$.
The kernel $\omega(\sigma)$ degrades to the impulse response $\delta$ when $\sigma$ equals to 0.

Then we introduce the new path function according to the above baseline input.
IG employs linear interpolation function as the path function.
However, the linear interpolated images show artifacts and do not follow the prior distribution of natural images.
To address this problem, we propose the \emph{progressive blurring path function} $\gamma_{\mathrm{pb}}$, which achieves a smooth transformation from $I'$ to $I$ through progressively changing the blur kernel:
\begin{equation}
    \gamma_{\mathrm{pb}}(\alpha)=\omega(\sigma - \alpha\sigma)\otimes I.
    \label{eq:PathFunction}
\end{equation}
Obviously, we have $\gamma_{\mathrm{pb}}(0)=I'$ and $\gamma_{\mathrm{pb}}(1)=I$.

In practice, we calculate the gradients at points sampled uniformly along the path and then approximate Eq~\eqref{eq:PathIntegratedGradient} with a summation:
\begin{align}
    &\mathsf{LAM}_{F,D}(\gamma_{\mathrm{pb}})^{approx}_i:=
    \label{eq:PathIntegratedGradientApprox}
    \\
    &\sum_{k=1}^m \frac{\partial D(F(\gamma_{\mathrm{pb}}(\frac{k}{m}))))}{\partial\gamma_{\mathrm{pb}}(\frac{k}{m})_i}\cdot\Big(\gamma_{\mathrm{pb}}(\frac{k}{m})-\gamma_{\mathrm{pb}}(\frac{k+1}{m})\Big)_i,\notag
\end{align}
where $m$ is the number of steps in the approximation of the integral.
Here, we approximate $\frac{\partial\gamma_{\mathrm{pb}}(\alpha)}{\partial\alpha}$ by calculating the difference after discretization.
The calculation of gradients in Eq~\eqref{eq:PathIntegratedGradientApprox} can be directly implemented using computational graph frameworks, \eg, PyTorch \cite{pytorch} and Tensorflow \cite{tensorflow}.
Experimentally, a step number of 100 is enough to approximate the integral.
One can check the accuracy of the approximation according to the proposition \cite{sundararajan2017axiomatic,friedman2004paths} that $\sum_i \mathsf{LAM}_{F,D}(\gamma_{\mathrm{pb}})_i=D(F(I))-D(F(I'))$.

\begin{figure}[t]
    \centering
    \includegraphics[width=0.95\linewidth]{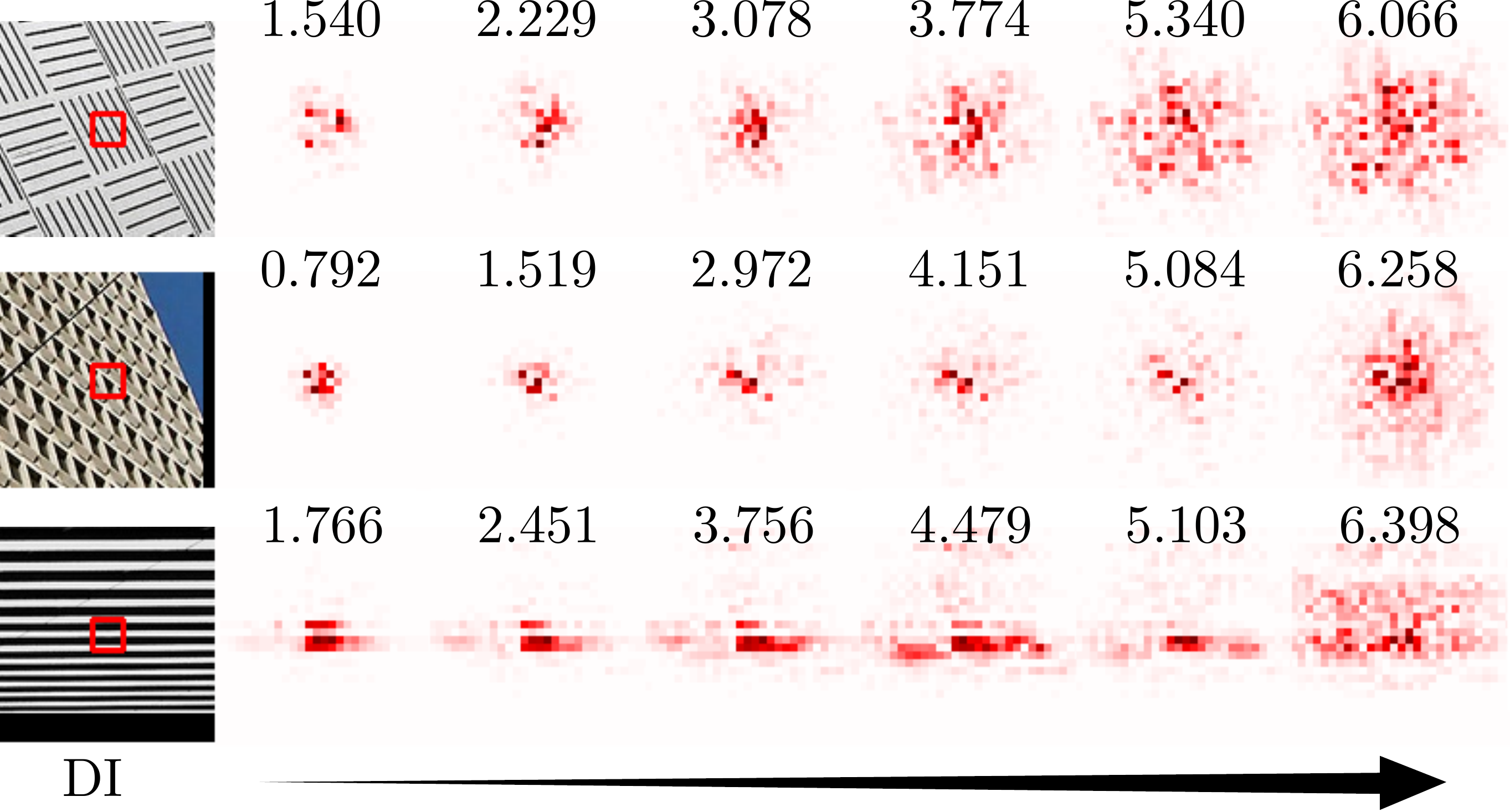}
    \vspace{-3mm}
    \caption{Attribution results with different diffusion index (DI). The DI reflects the range of involved pixels. A higher DI represents a wider range of attention.}
    \label{fig:gini}
    \vspace{-6mm}
\end{figure}

\subsection{Discussion}
\label{sec:mtd:disc}
\noindent\textbf{Why using integrated gradient.}\quad
It is counter-intuitive to use the path integrated gradients instead of the vanilla gradients, as the latter one directly indicates the direction of the maximum increase of the detected feature existence.
However, the vanilla gradients usually have small magnitudes and fail to indicate important features.
This is also called the ``gradient saturation''.
We illustrate the saturation by plotting the curve of $D(F(\gamma(\alpha)))$ along with the change of $\alpha$.
As shown in \figurename~\ref{fig:bsl&path}.f, saturation can be triggered when the curve shows a flat trend, where a small shift of $I$ along with the gradients does not significantly change the value of $D(F(I))$.
On the contrary, the path integrated gradients shows how $D(F(I))$ changes from small to large and its attribution map indicates the pixels that contribute most significantly to $D(F(I))$ when changing alone the path.
We illustrate this process by decomposing Eq~\eqref{eq:PathIntegratedGradient} into two parts: the gradients at the interpolated images and the gradients of $\gamma(\alpha)$, where the latter one could be viewed as weights for the gradients at different interpolations.
As shown in \figurename~\ref{fig:bsl&path}.f, the change of $D(F(\gamma_{\mathrm{pb}(\alpha)}))$ is not linear.
It experiences a sharp increase in the range marked by the green dashed line.
Meanwhile, the magnitudes of $\frac{\partial\gamma_{\mathrm{pb}}(\alpha)}{\partial\alpha}$ also have large values in this range, which makes these gradients provide greater contribution to the final attribution map.
We also show the curve of the cumulative gradients over $\alpha$ in \figurename~\ref{fig:bsl&path}.h.
Notice that high magnitude gradients accumulate at the same range where the $D(F(\gamma_{\mathrm{pb}(\alpha)}))$ increases sharply and also where the magnitudes of $\frac{\partial\gamma_{\mathrm{pb}}(\alpha)}{\partial\alpha}$ have large values.
These experiments show the rationality of using the path integral gradient instead of the vanilla gradient.

\noindent\textbf{The choice of baseline and path function.}\quad
Although Sundararajan \etal \cite{sundararajan2017axiomatic} suggest to use black image as baseline and linear interpolation as the path function, Sturmfels \etal \cite{sturmfels2020visualizing} argue that it may not be the best choice.
We experimentally show its disadvantages in \figurename~\ref{fig:bsl&path}.
As can be seen, the linear interpolated images of black baseline image does not present the ``absence'' of important features, although the feature detection output of the baseline image is reduced.
It reduces $D(F(I))$ by reducing the intensity of all the pixels, resulting in saturation for the gradients of all these interpolations.
On the other hand, $\frac{\partial\gamma_{\mathrm{lin}}(\alpha)}{\partial\alpha}$ is equal to $I-I'$ for all $\alpha$ in linear path function, which provides all gradients the same weights.
For these reasons, the black baseline image and linear path function are not suitable for interpreting the SR networks.
Alternative choices are proposed in this work for SR networks.
We use a blurred image as baseline to represent the missing high-frequency components.
The the progressively blurring function is presented as a natural choice of path function. 
With the mathematical characteristics of path integral gradients, the proposed method can provide reasonable attribution results for SR networks.

\begin{figure*}
    \centering
    \includegraphics[width=\linewidth]{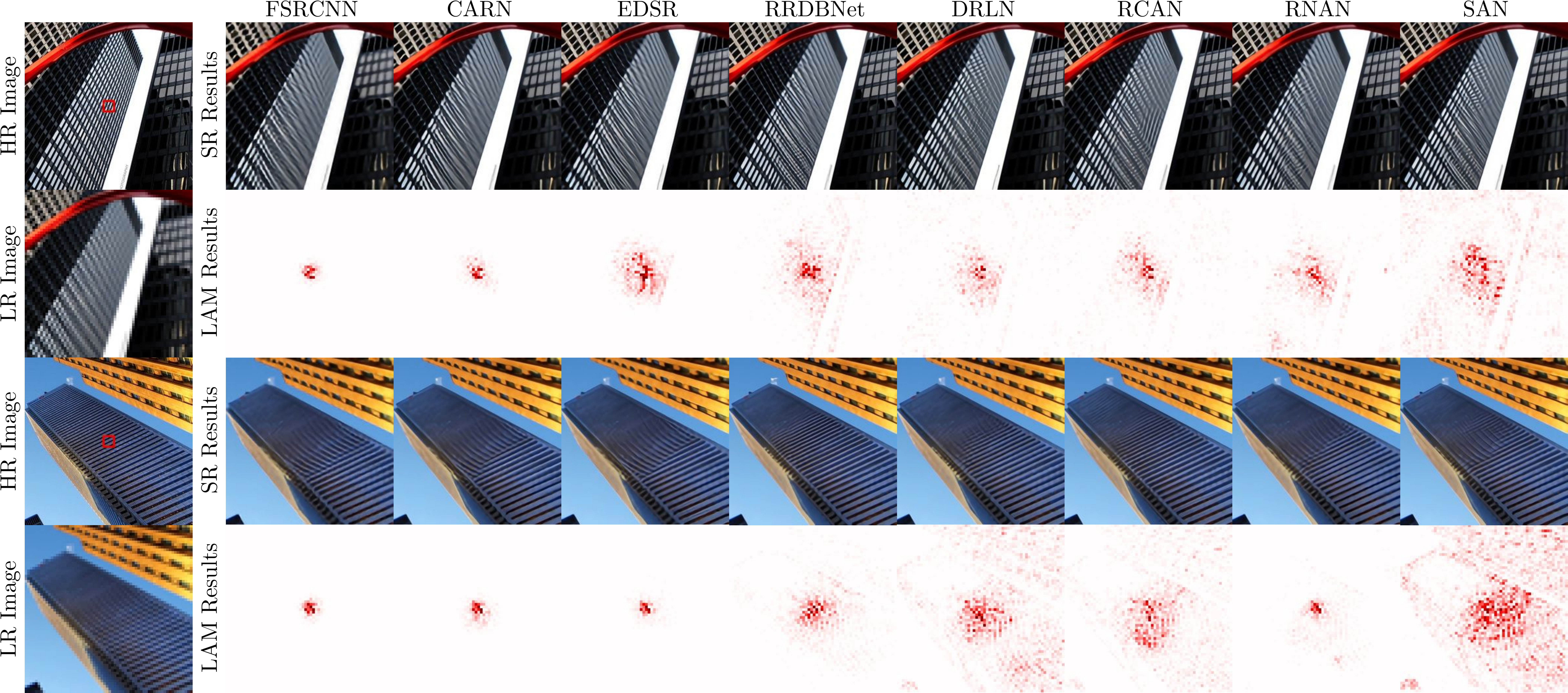}
    \vspace{-7mm}
    \caption{Comparison of the SR results and LAM attribution results of different SR networks. The LAM results visualize the importance of different pixel \wrt the SR results.}
    \label{fig:lam_results}
    \vspace{-3mm}
\end{figure*}

\begin{figure*}
    \centering
    \includegraphics[width=\linewidth]{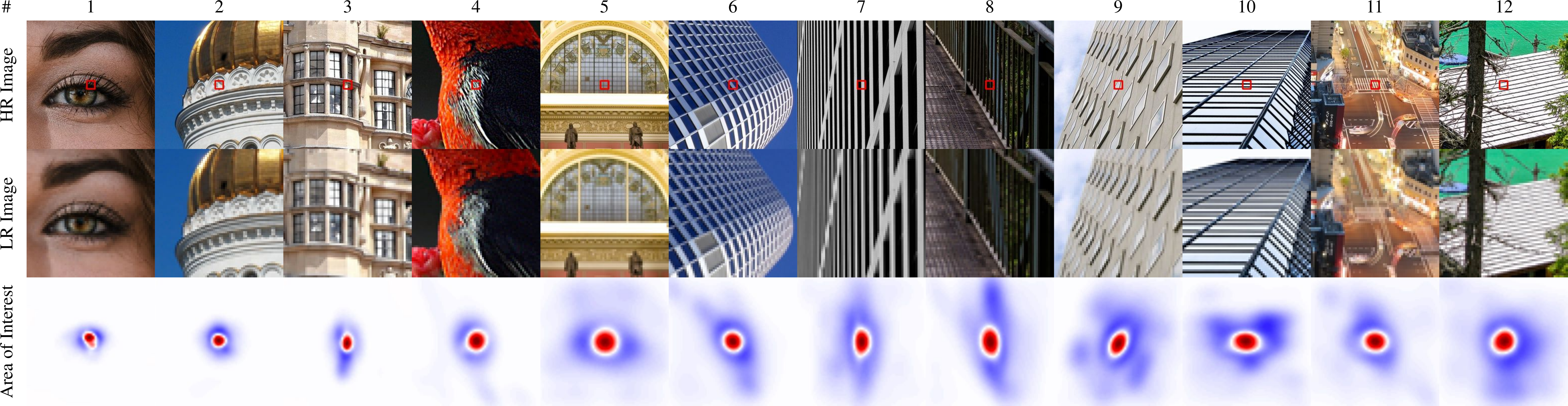}
    \vspace{-7mm}
    \caption{The heat maps exhibit the area of interest for different SR networks. The pixels with \textcolor{red}{red} color are noticed by almost all SR networks while the areas marked with \textcolor{blue}{blue} represents the differences between the SR networks with large LAM interest areas and those with small interest areas.}
    \label{fig:area_of_interest}
    \vspace{-6mm}
\end{figure*}

\section{Experiments}
\subsection{Collection of Test Set and Visualization Method}
Following the principle of interpreting hard cases, we collect 150 images that are challenging for SR networks as the test set for the following analysis.
We first sample more than 30,000 sub-images of size $256\times256$ from DIV2K validation set \cite{div2k} and Urban100 \cite{huang2015single} and then obtain their $\times4$ SR results using different SR networks.
We select the sub-images that have low average PSNR performance and high variance between different SR networks and then manually remove images with duplicate and unidentifiable content.
In practice, we only perform attribution analysis to the $16\times16$ local patch in the center of the image, so we manually adjust the image to make the content in the center meaningful.
In our test set, the average PSNR value is only 20.87dB.
Compared with 28.59dB and 24.12dB (the average PSNR values of DIV2K validation set and Urban100), our test images are challenging for SR networks.
When selecting metrics for quantitative evaluation, we follow the suggestion of Gu \etal \cite{gu2020pipal} and employ both PSNR and the LPIPS perceptual similarity \cite{zhang2018unreasonable}.
For the visualization of the attribution results, we first normalize the maps to the range $[-1, 1]$ and then take the absolute values.
Although the normalization only keeps the relative values without signs, we contend that such properties will not affect the perception of the output visualization.
Similar visualization methods are also used in some previous work \cite{adebayo2018sanity}.
In the visualization results, a darker pixel (larger intensity) indicates a larger influence \wrt the SR results.
In the following texts, the heat maps of distributions are obtained using kernel density estimation \cite{kde1}.

\subsection{Diffusion Index for Quantitative Analysis}
As stated above, LAM highlights the pixels which have the greatest impact to the SR results.
Theoretically, for the same local patch, if the LAM map involves more pixels or a larger range, it can be considered that the SR network has utilized information from more pixels.
For quantitative analysis, we employ the Gini coefficient \cite{Gini} to indicate the range of involved pixels, denoted as
\begin{equation}
    G=\frac{\sum_{i=1}^n\sum_{j=1}^n|g_i-g_j|}{2n^2\bar{g}},
\end{equation}
where $g_i$ represents the absolute value of the $i$th dimension of the attribution map, $\bar{g}$ is the mean value and $G\in[0,1]$.
Gini coefficient is originally a measure of statistical dispersion intended to represent the income inequality.
In our case, the inequality of pixels' contribution to the attribution result also reflects the range of involved pixels.
If a few pixels contribute a large proportion in the total attribution result, its Gini coefficient is relatively high, otherwise, a low Gini coefficient indicates the attribution result involves more pixels.
In practice, we rescale the Gini coefficient and propose the diffusion index (DI) to facilitate analysis, denoted as
\begin{equation}
    \mathrm{DI} = (1-G)\times100.
\end{equation}
Notice that in this setting, a larger DI indicates more pixels are involved.
We illustrate the Gini and diffusion coefficient in \figurename~\ref{fig:gini} with examples.

\begin{figure*}[t]
    \centering
    \includegraphics[width=\linewidth]{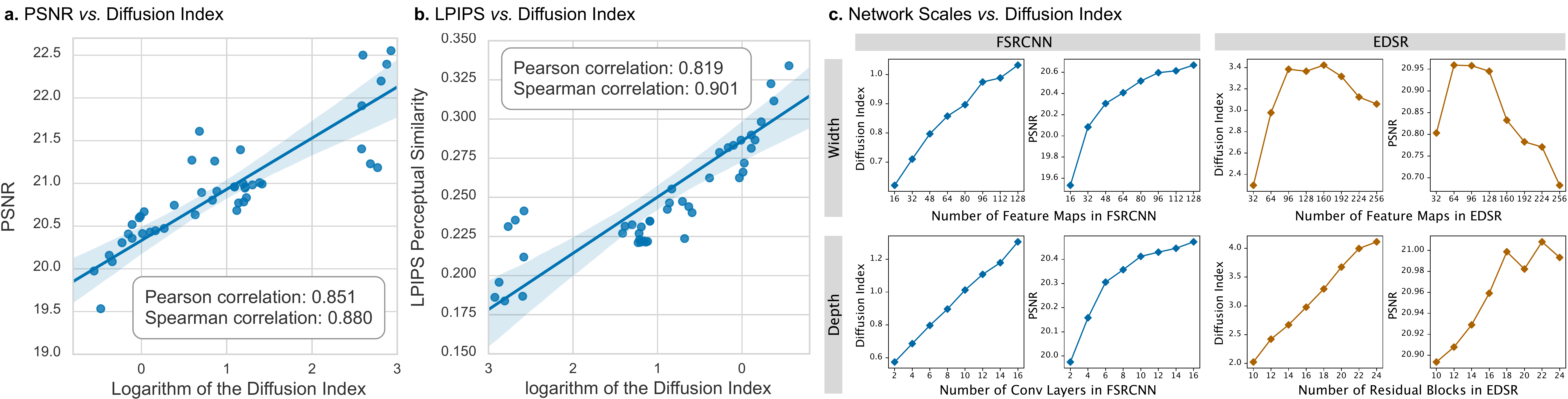}
    \vspace{-7mm}
    \caption{Key results related with DI. (\textbf{a}) the linear fitting results of DI \vs PSNR performance. (\textbf{b}) the linear fitting results of DI \vs LPIPS perceptual similarity performance. (\textbf{c}) DI values with different network scales for FSRCNN and EDSR.}
    \label{fig:overview}
    \vspace{-4mm}
\end{figure*}

\subsection{Attribution Results}
We show some LAM results of different SR networks in \figurename~\ref{fig:lam_results}.
We can have the following observations.
First, the early network FSRCNN \cite{fsrcnn2016} and shallow network CARN \cite{carn2018} have relatively small receptive fields and can only perform SR based on limited surrounding pixels.
Deep residual networks, \ie, EDSR \cite{edsr2017}, RRDBNet \cite{wang2018esrgan}, RCAN \cite{rcan2018} and DRLN \cite{drln2020}, have deeper architectures and bigger receptive fields.
Their LAM results show that these networks are interested in a wider range of pixels in the LR images with similar patterns, \eg, the regular stripes that appear on skyscrapers.
Although the textures in these areas are heavily aliased in the LR image, which may mislead the SR process, some networks still reconstruct accurate textures and they, according to the LAM results, take notice of a wider range of unaliased areas.
Second, non-local and channel-wise attention modules are introduced to SR networks in recent years, expecting to utilize long-term and global information to assist the SR.
Representative network designs include RNAN \cite{nonResidual}, RCAN and SAN \cite{san2019}.
The proposed LAM method can also be applied to diagnose and visualize that whether global and long-term information is used by these networks.
As can be observed from \figurename~\ref{fig:lam_results}, they all extract information from non-local pixels.
This experiment demonstrates the value of LAM as a research aid.
Third, it’s worth noting that some networks also take a wider range of pixels into account but still reconstruct wrong textures, \eg, SAN and DRLN in the second example.
In these models, the information from more pixels is not effectively used for accurate reconstruction.

We then visualize the similarities and differences of LAM results for different SR networks in \figurename~\ref{fig:area_of_interest}.
The heat maps in the third row show the areas of interest for different SR networks.
The information carried in the red areas can be used for the most preliminary level of SR, while the blue areas show the potential informative areas that can further improve SR.
How to extract and use the information in blue areas is the key factor for a SR network to distinguish with others.
To empirically analyze the patterns of information utilization, we find that they are more likely to notice areas with self-similar properties and regular textures.
For instance, in the 6th, 7th, and 8th examples of \figurename~\ref{fig:area_of_interest}, the attention of the SR networks is distributed along the direction of texture extension; and in the 9th example, the SR networks notice the similar-shaped windows around the target patch.
We include more results in Appendix~\ref{sec:apd:more}.

\begin{table}[t]
    \centering
    \resizebox{1.0\linewidth}{!}{
    \begin{tabular}{lcccp{4cm}}
    \hline
        Model & Recpt. Field & PSNR & DI & Remark \\
        \hline
        FSRCNN & 17$\times$17 & 20.30 & 0.797 & Fully convolution network. \\
        CARN & 45$\times$45 & 21.27 & 1.807 & Residual network. \\
        EDSR & 75$\times$75 & 20.96 & 2.977 & Residual network.\\
        MSRN & 107$\times$107 & 21.39 & 3.194 & Residual network.\\
        RRDBNet & 703$\times$703 & 20.96 & 13.417 & Residual network.\\
        \hdashline
        IMDN & global & 21.23 & 14.643 & Global pooling. \\
        RFDN & global & 21.40 & 13.208 & Global pooling. \\
        RCAN & global & 22.20 & 16.596 & Global pooling. \\
        RNAN & global & 21.91 & 13.243 & Non-local attention. \\
        SAN & global & 22.55 & 18.642 & Non-local attention. \\
    \hline
    \end{tabular}
    }
    \vspace{-2mm}
    \caption{\small Comparison of the receptive fields, PSNR and DI for some representative SR networks. For computing receptive fields, we follow the guidance provided by Araujo \etal \cite{araujo2019computing}.}
    \label{tab:receptivefield}
    \vspace{-6mm}
\end{table}

\subsection{Exploration with LAM}
In this section, we use LAM as a tool to explore SR networks.
We study the relationship between the diffusion index of LAM results and several indicators of SR networks, \eg, performance, scales and receptive fields.
In our experiment, totally 48 SR networks are collected for analysis, consisting of 18 networks from the literature and 30 networks with different network scales.
For networks from the literature, we select:
SRCNN \cite{srcnn2014}, 
FSRCNN, 
EDSR, 
LapSRN \cite{lapsrn2017}, 
SRResNet \cite{srgan2017}, 
DDBPN \cite{ddbpn2018}, 
RDN \cite{rdn2018}, 
MSRN \cite{msrn2018}, 
RCAN, 
RRDBNet, 
CARN, 
SAN, 
IMDN \cite{imdn2019}, 
RNAN \cite{nonResidual}, 
RFDN \cite{rfdn2020}, 
PAN \cite{pan2020}, 
DRLN, and 
CSNLN \cite{crossNonlocal}. 
For networks with different network scales, we collect networks for two representative architectures, \ie, FSRCNN \cite{fsrcnn2016} as a representative of the fully convolution networks and EDSR \cite{edsr2017} as a representative of the residual networks.
For FSRCNN, the range of width is selected to be $[16, 128]$ and the range of depth is $[2,16]$, with 15 models in total. 
For EDSR, the range of width is selected to be $[32, 256]$ and the range of residual blocks is $[10,24]$, with 15 models in total. 
All the models are trained with the same setting.
More details about model collection can be found in Appendix~\ref{sec:apd:models}.

\noindent\textbf{Diffusion Index \vs Network Performances.}\quad
The question that we are most interested in is ``can we get better SR performance by utilizing information from more pixels?''
We establish this relationship with linear regression fitting, where the x-axis represents the log number of DI and the y-axis represents the SR performance.
We illustrate the results in \figurename~\ref{fig:overview}.a and \figurename~\ref{fig:overview}.b.
The high pearson correlation and spearman correlation (both larger than 0.8) indicate that the extraction of information from a wider range of pixels and SR performance are highly correlated, the two-tailed p-test results for both experiments are lower than $1\times10^{-12}$.
Finding ways to capture the benefits from a wider range of pixels is an important direction for future work.

\noindent\textbf{Diffusion Index \vs Receptive Field.}\quad
A result of network deepening is the increase of the receptive field, and some networks also propose to increase the receptive field on purpose to promote SR.
We are curious whether the increase in the receptive field directly promotes the use of information from a wider range of pixels.
In \tablename~\ref{tab:receptivefield} we show the receptive fields, PSNR and DI performances for some representative SR networks.
As the number of convolution layers increases, the receptive field steadily increases, and the area and pixels that the network can perceive also increase.
In addition to adding more convolution layers, channel-wise attention with global pooling layers and non-local operations are also introduced, which can theoretically increase the receptive field to cover the whole image.
Compared with RRDBNet (whose receptive field is already larger than the size of the test images), IMDN, RFDN and RNAN with the above global operations do not significantly utilize the information from more pixels.
However, they are more efficient in the use of information, which is reflected in their better PSNR performance.
Besides, RCAN employs not only global operations but also more than 400 convolution layers that can support a very large effective receptive field, and SAN employs multiple non-local operations and also global-wised attentions.
They all achieve higher DI values and also better performances.
We argue that the receptive field of the existing networks is large enough, and the effect of simply increasing the receptive field is limited.
How to effectively utilize the information within the receptive area pixels is critical.

\noindent\textbf{Diffusion Index \vs Network Scales.}\quad
SR networks have grown deeper and wider in the quest for higher reconstruction accuracy. 
We now investigate the relationship between the attribution results and different SR network scales (width and depth).
The results are shown in \figurename~\ref{fig:overview}.c.
As one can observe, with the increase of both depth and width, FSRCNN is able to extract information from more pixels and also achieves better PSNR performance.
For EDSR, the increase of the number of convolution layers directly increases the receptive field and leads to better DI values and PSNR performance, however, the increase of the number of feature maps in each layer does not achieve the corresponding improvement.
This is because, under the same training condition, networks with larger parameters are difficult to optimize effectively.

\begin{figure}[t]
    \centering
    \includegraphics[width=\linewidth]{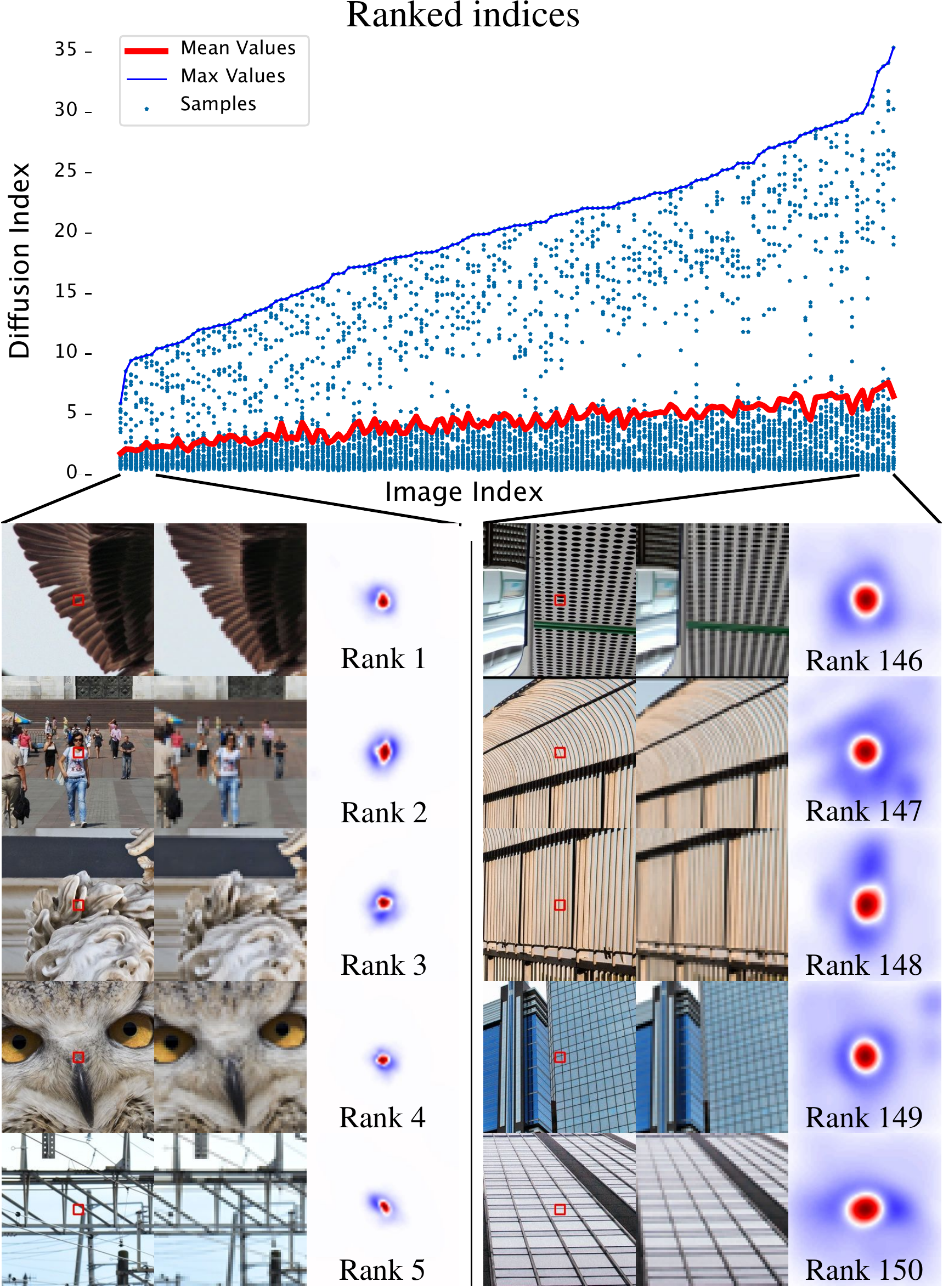}
    \vspace{-4mm}
    \caption{Relationship of DI and image contents. The scatter plot shows the distribution of the DI values for different images and SR networks. The left bottom images with low rank indices show the images with narrow area of interest, and the right bottom images with high rank indices show the images with large area of interest.}
    \label{fig:bigger_lam}
    \vspace{-6mm}
\end{figure}

\noindent\textbf{Diffusion Index \vs Image Content.}\quad
We have also observed significant differences in LAM results on images with different content.
We first sort the images with highest DI values and the results are shown in \figurename~\ref{fig:bigger_lam}.
As can be observed, the network's perception of different image contents is also different.
In some images, even for the SR network with a large receptive field and good learning capacity, the area of noticing is still narrow.
It indicates that these networks believe the semantics or features in a wider area have little help to the SR of the current patch.
On the contrary, for some other images, these networks can extract features from a wider surrounding area.
We illustrate these two image categories in \figurename~\ref{fig:bigger_lam}.
As can be observed, images with regular stripe and grid textures are very likely to be detected by SR networks, while the complex high-level semantics, such as human and animal, can not be effectively used.
The results of LAM actually point out that for SR networks, the extracted and used abstract semantics are different from what people usually understand.
It is an open question that whether the use of complex high-level semantics can help SR.

\section{Conclusion}
In this paper, we propose local attribution map (LAM) to visualize and understand SR networks.
We experimentally demonstrate the potential of LAM as a research tool and reveal some interesting conclusions about what information is used by SR networks and how this information affect performance.
Our work opens new directions for designing SR networks and interpreting low-level vision deep models.

{\small
\bibliographystyle{ieee_fullname}
\bibliography{egbib}
}

\appendix

\section*{Appendix}

\section{Review of Attribution Methods}
\label{sec:apd:overview}
In this section, we provide a review of attribution methods in the literature that are used for interpreting classification networks.
We also discuss their relationship with the interpretation of super-resolution (SR) networks.
Given an input image $I\in\mathbb{R}^d$ and a model $S: \mathbb{R}^d\mapsto\mathbb{R}$ that outputs the probability of $I$ belongs to a certain class, an attribution method provides attribution maps $\mathsf{Attr}_S:\mathbb{R}^d\mapsto\mathbb{R}^d$ for $S$ that are of the same size as the inputs.
Each dimension of these attribution maps corresponds to the ``relevance'' or ``importance'' of that dimension to the final output, which is often a class-specific score in classification networks.

\noindent\textbf{Gradient \wrt $I$.}\quad
This method employs the gradient of the predicted probability \wrt to the input $I$ \cite{simonyan2013deep,baehrens2010explain}.
\begin{equation}
    \mathsf{Grad}_{S}(I)=\frac{\partial S(I)}{\partial I}
\end{equation}
However, the vanilla gradient method suffers from the ``saturation'' problem that the magnitude of this gradient tends to be small.
A little movement toward the direction of the gradient will not change the predicted probability significantly \cite{sturmfels2020visualizing}.
In Sec~\ref{sec:mtd:disc}, we show that for the interpretation of SR networks, the ``saturation'' problem also exist.
Thus the vanilla gradient method is not appropriate for interpreting SR networks.

\noindent\textbf{The element-wise product of Gradient and the input.}\quad
This method was proposed to address the saturation problem and reduce visual diffusion \cite{shrikumar2016not}, denoted as
\begin{equation}
    \mathsf{Grad\odot I}_{S}(I)=I\odot\frac{\partial S(I)}{\partial I}.
\end{equation}
Ancona \etal \cite{ancona2018towards} show that, for a network with only ReLU activation function and no additive biases, this input gradient product is equivalent to DeepLift \cite{shrikumar2016not}, and $\epsilon$-LRP \cite{bach2015pixel}.
For the interpretation of SR networks, the pixel intensity should not be part of the attribution as the textures and edges may not change when the pixel intensity changes.
Directly calculate the product of the input intensity and the gradient will introduce interference factors.

\noindent\textbf{Guided Backpropagation (GBP).}\quad
This method specifies a change in how to calculate gradients for ReLU activations.
Let $\{f^l,f^{l-1},\dots,f^0\}$ be the feature maps obtained during the forward process by a deep neural network $S$, and $\{r^l,r^{l-1},\dots,r^0\}$ be the representation obtained during the backward process.
Springenberg \etal \cite{springenberg2015striving} propose GBP that aims to zero out negative gradients during the computation of $r$.
The map is computed as:
\begin{equation}
    r^l=1_{r^{l+1}>0}1_{f^l>0}r^{l+1},
\end{equation}
where $1_{r^{l+1}>0}$ represents keeping only the positive gradients and $1_{f^l>0}$ indicates keeping only the positive activations.
The usage scenarios of this method are relatively limited.
For residual networks that are widely used in SR, this method is not valid.

\noindent\textbf{Integrated Gradients (IG).}\quad
Most relevant to the method proposed in this paper, IG also employs path integration \cite{friedman2004paths}, but uses a black image as baseline image and linear interpolation as the path function.
IG is defined as:
\begin{equation}
    \mathsf{IG}_{S}(I)=(I-I')\times\int_0^1\frac{\partial S(I'+\alpha(I-I'))}{\partial I}d\alpha,
\end{equation}
where $I'$ is the baseline black image and $\alpha$ is the parameter of the interpolation.
In Sec~\ref{sec:mtd:disc}, we discuss the differences between the proposed local attribution maps for SR networks and IG.

\noindent\textbf{SmoothGrad and VarGrad.}\quad
SmoothGrad \cite{smilkov2017smoothgrad} and VarGrad \cite{adebayo2018local} are proposed to relieve the situation where the attribution graph is full of noise.
The SmoothGrad is defined as:
\begin{equation}
    \mathsf{SmoothGrad}_{S}(I)=\frac{1}{N}\sum_{i=1}^N\mathsf{Grad}_{S}(I+n_i),
\end{equation}
where $n_i$ are the noise vectors and $n_i\sim\mathcal{N}(0,\sigma^2)$ are sampled from a Gaussian distribution.
Similar to SmoothGrad, a variance analog of SmoothGrad can be defined as:
\begin{equation}
    \mathsf{VarGrad}_{S}(I)=\mathcal{V}(\mathsf{Grad}_{S}(I+n_i)),
\end{equation}
where $\mathcal{V}$ represents to the variance.
Seo \etal \cite{seo2018noise} theoretically analyze VarGrad showing that it is independent of the gradient, and captures higher order partial derivatives.
For SR networks, adding noise to the input is destructive to the output image \cite{qian2019trinity,gu2019blind}.
Thus both SmoothGrad and VarGrad can not be used to interpret SR networks.
On the other hand, SmoothGrad and VarGrad also face the challenge of gradient saturation.

\noindent\textbf{CAM, GradCAM and Guided GradCAM.}\quad
Different from the aforementioned gradient-based attribution methods, Class Activation Mapping (CAM) \cite{zhou2016learning} generates class activation maps using the global average pooling in convolution neural networks.
A CAM map for a particular category indicates the discriminative image regions used by the network to identify that category.
Combining gradient-based methods and CAM, Selvaraju \etal \cite{selvaraju2017grad} further propose GradCAM that corresponds to the gradient of the class score \wrt the feature map of the last convolution unit.
For pixel level granularity, GradCAM can be combined with Guided Backpropagation through an element-wise product.
Since CAM is specially designed for high-level vision networks with global pooling layers, it cannot be easily adapted to low-level vision models such as SR networks.

\noindent\textbf{Perturbation-Based Methods.}\quad
Different from the above works that require the mathematical details of the model, there are works that treat deep models as black-boxes.
These methods usually localize the discriminative image regions by performing perturbation to the input.
For instance, Fong and Vedaldi \cite{fong2017interpretable} propose to explain neural networks that are based on learning the minimal deletion to an image that changes the model prediction.
Similar to SmoothGrad and VarGrad, the sensitivity of SR networks to disturbances and perturbation makes it difficult to use these approach to explain.

\section{Collection of Models}
\label{sec:apd:models}
In this section, we first describe the training settings in our experiments and then briefly review the used SR networks.
We use DIV2K training set \cite{div2k} for training and the size of LR image is $64\times64$.
For optimization, we use Adam \cite{kingma2014adam} with the default settings that $\beta_1=0.9$ and $\beta_2=0.999$.
The learning rate is initialized as $1\times10^{-4}$ and decayed linearly at every $2\times10^5$ updates.
The size of minibatch is set to 16.
We next briefly review the used SR networks.

\noindent\textbf{Early methods with fully convolutional architectures.}\quad
These methods include SRCNN \cite{srcnn2014}, FSRCNN \cite{fsrcnn2016} and ESPCN \cite{espcn2016}.
What they have in common is that they only use stacked convolution layers without residual or other deep modules.
SRCNN is the first deep SR network that consists of only three convolution layers without upsampling layer -- it takes the bicubic interpolation result as input.
FSRCNN consists of eight convolution layers and uses deconvolution layer as the upsampling layer.
In ESPCN, pixel shuffle is used innovatively as an upsampling operation, and this operation is used on a large scale by subsequent SR networks.
In addition to the above networks, DDBPN \cite{ddbpn2018} and LapSRN \cite{lapsrn2017} are also in the form of fully convolution networks with different convolution strategies.
LapSRN is a network with progressive upsampling operations that super-resolves low-resolution images in a coarse-to-fine laplacian pyramid framework.
DDBPN exploits iterative up- and downsampling layers, aiming at providing an error feedback mechanism for projection errors at each stage.

\noindent\textbf{Networks with residual and dense connections.}\quad
These methods date back to SRResNet \cite{srgan2017} that first introduce residual connections \cite{he2016deep} to deep SR networks.
Some methods are proposed to improve the residual structure such as EDSR \cite{edsr2017}, CARN \cite{carn2018} and MSRN \cite{msrn2018}.
Spatial feature transformation blocks are also introduced to SR networks \cite{sftgan2018,gu2019blind} to achieve interactive SR.
Inspired by dense connection network \cite{huang2017densely}, RDN \cite{rdn2018} and SRDenseNet \cite{srdense2017} with dense architecture was proposed.
Combining residual blocks and dense connections, residual-in-residual dense net (RRBDNet) \cite{wang2018esrgan} was proposed.
Recently, DRLN \cite{drln2020} employs cascading residual on the residual structure to allow the flow of low-frequency information to focus on learning high and mid-level features.

\noindent\textbf{Networks with attention modules.}\quad
In addition to innovations in various short connections, attention modules are also used to improve the performance of SR networks.
Zhang \etal \cite{rcan2018} propose channel attention that compute attention weights \wrt the whole channel.
Zhao \etal \cite{pan2020} propose pixel attention that compute attention weights using $1\times1$ convolution for each pixel.
Non-local operation is also introduced in the form of attention module in \cite{nonResidual,nonRecurrent}.
SAN \cite{san2019} utilizes both non-local attention modules and second-order channel attention.
Recently, CSNLN \cite{crossNonlocal} employs cross-scale non-Local attention module with integration into a recurrent neural network to learn cross-scale feature correlation.


\section{More Results}
\label{sec:apd:more}
\begin{figure*}
    \centering
    \includegraphics[width=\linewidth]{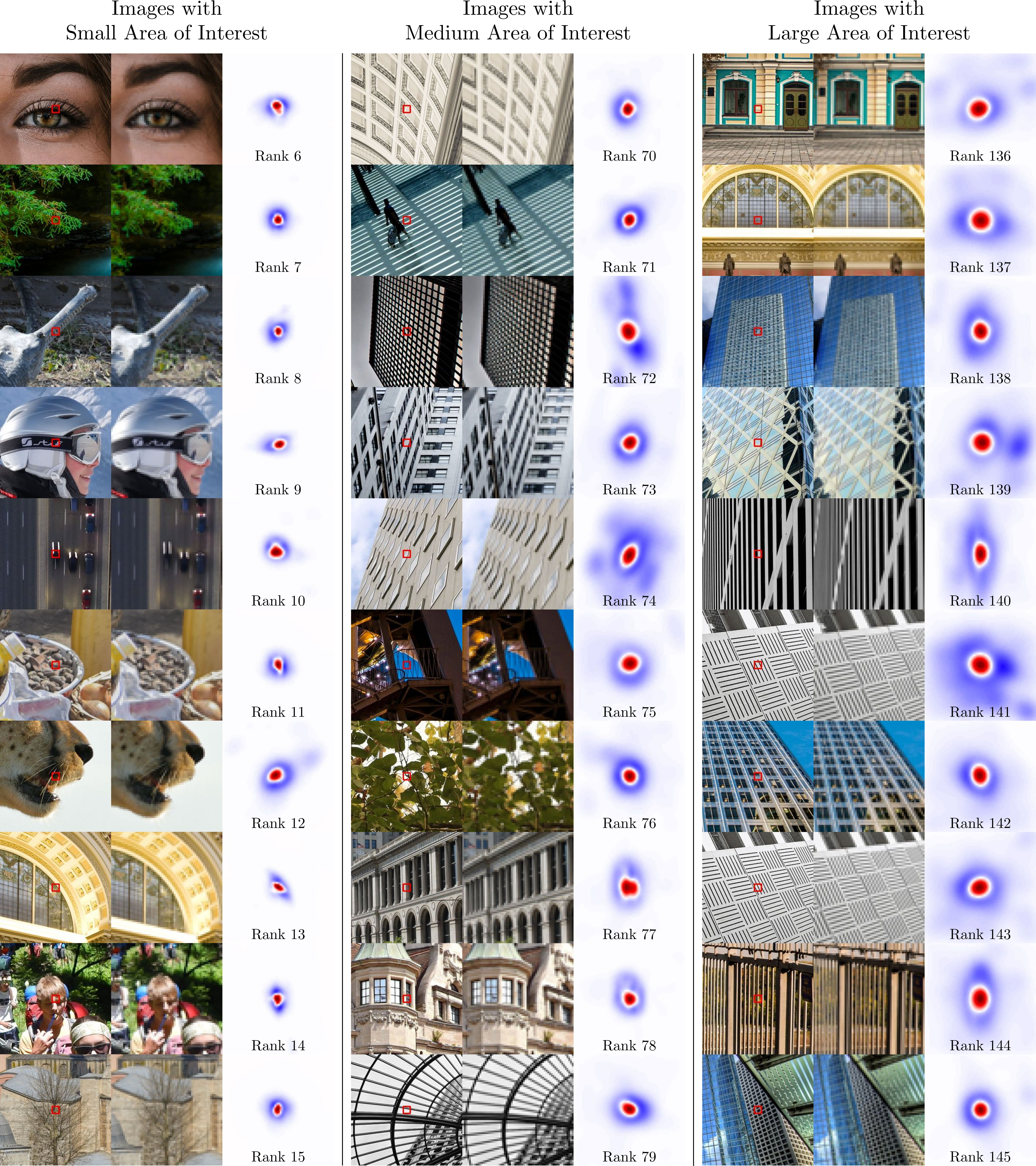}
    \caption{The heat maps exhibit the area of interest for different SR networks. The pixels with \textcolor{red}{red} color are noticed by almost all SR networks while the areas marked with \textcolor{blue}{blue} represents the differences between the SR networks with large LAM interest areas and those with small interest areas. The rank indices indicate the ranking order of the largest diffusion index of images' lam results.}
    \label{fig:apd:area_of_interest}
\end{figure*}

\begin{figure*}
    \centering
    \includegraphics[width=\linewidth]{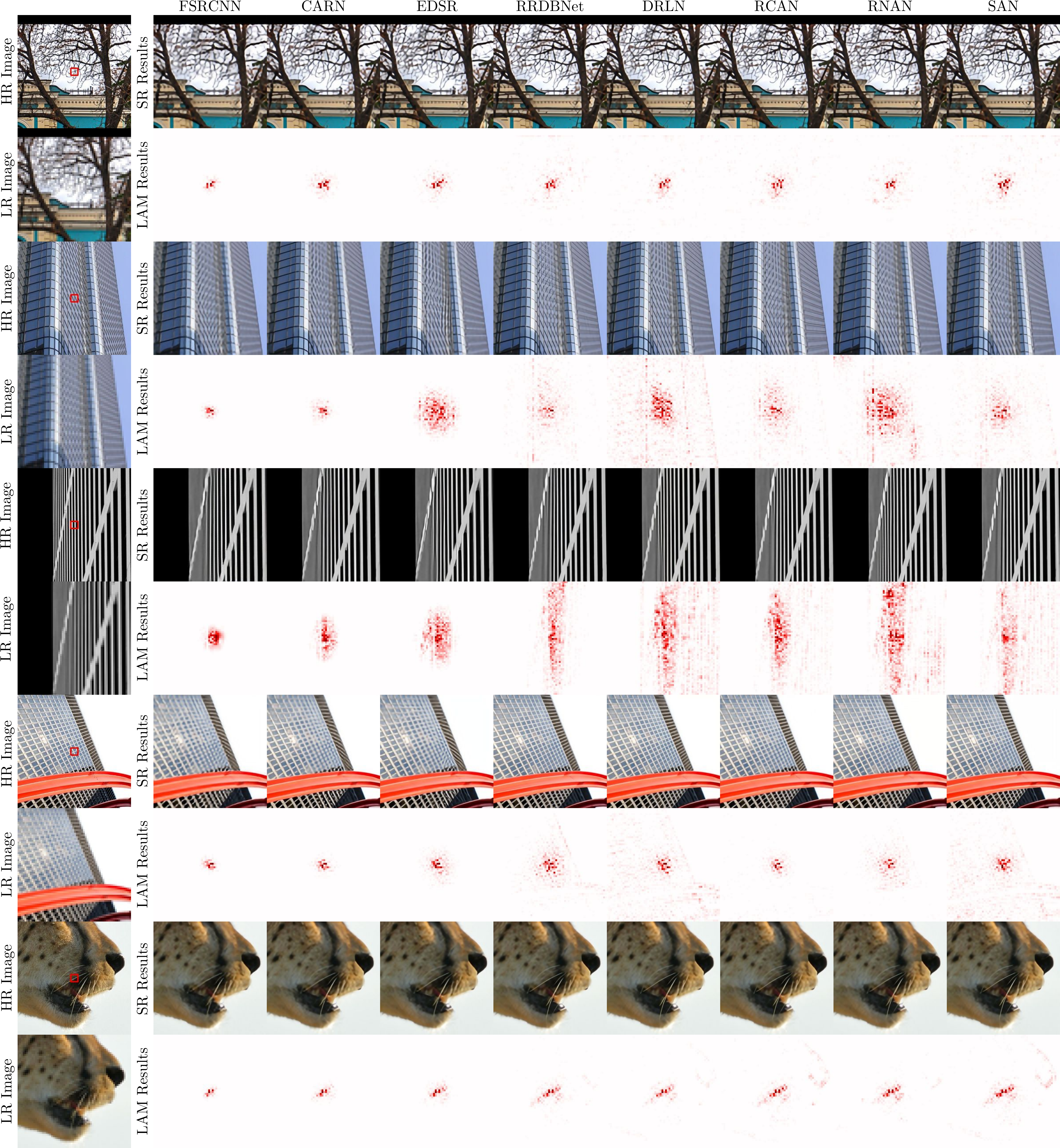}
    \caption{Comparison of the SR results and LAM attribution results of different SR networks. The LAM results visualize the importance of different pixel \wrt the SR results.}
    \label{fig:apd:lam_results_1}
\end{figure*}

\begin{figure*}
    \centering
    \includegraphics[width=\linewidth]{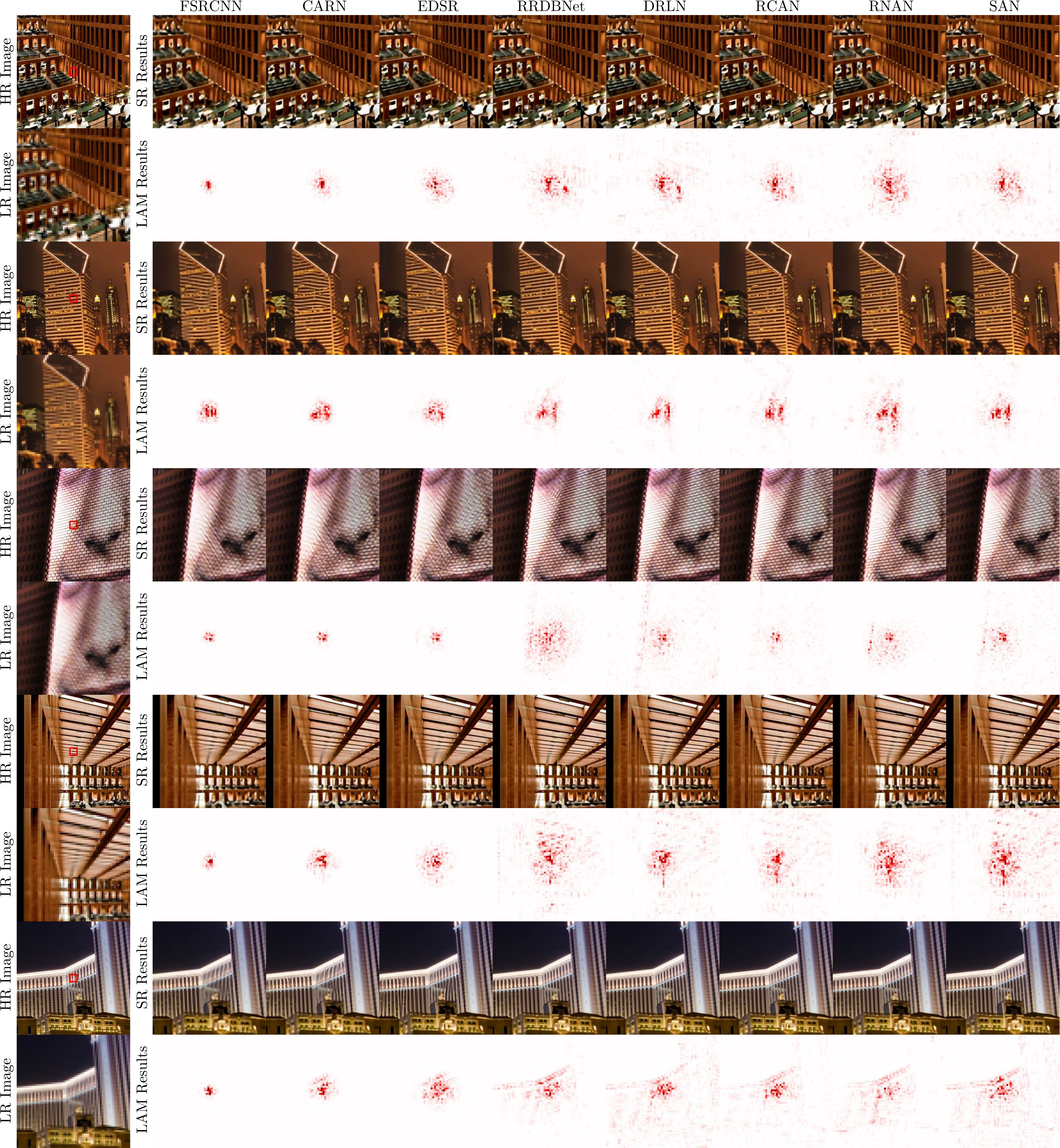}
    \caption{Comparison of the SR results and LAM attribution results of different SR networks. The LAM results visualize the importance of different pixel \wrt the SR results.}
    \label{fig:apd:lam_results_2}
\end{figure*}

\begin{figure*}
    \centering
    \includegraphics[width=\linewidth]{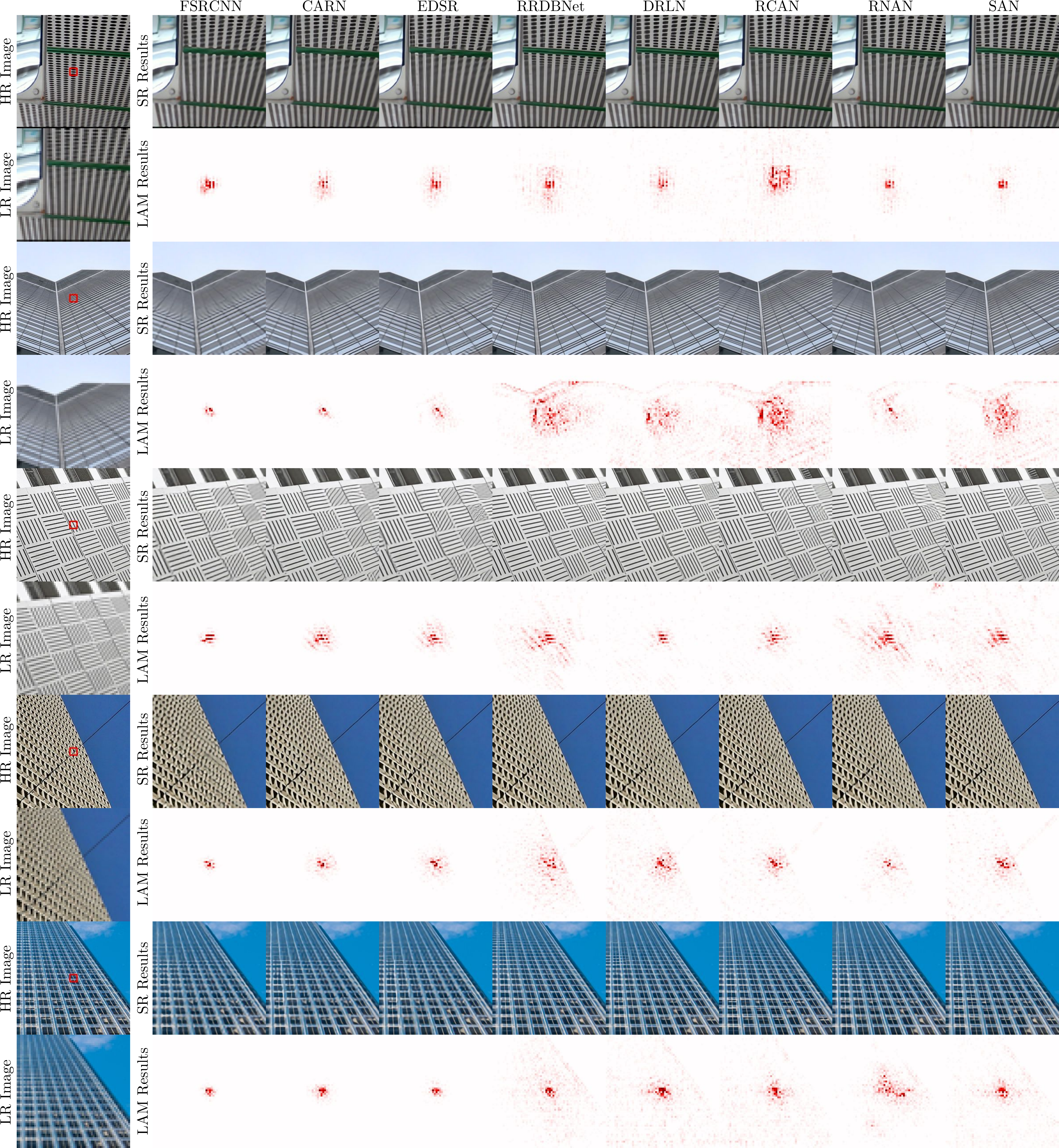}
    \caption{Comparison of the SR results and LAM attribution results of different SR networks. The LAM results visualize the importance of different pixel \wrt the SR results.}
    \label{fig:apd:lam_results_3}
\end{figure*}

\begin{figure*}
    \centering
    \includegraphics[width=\linewidth]{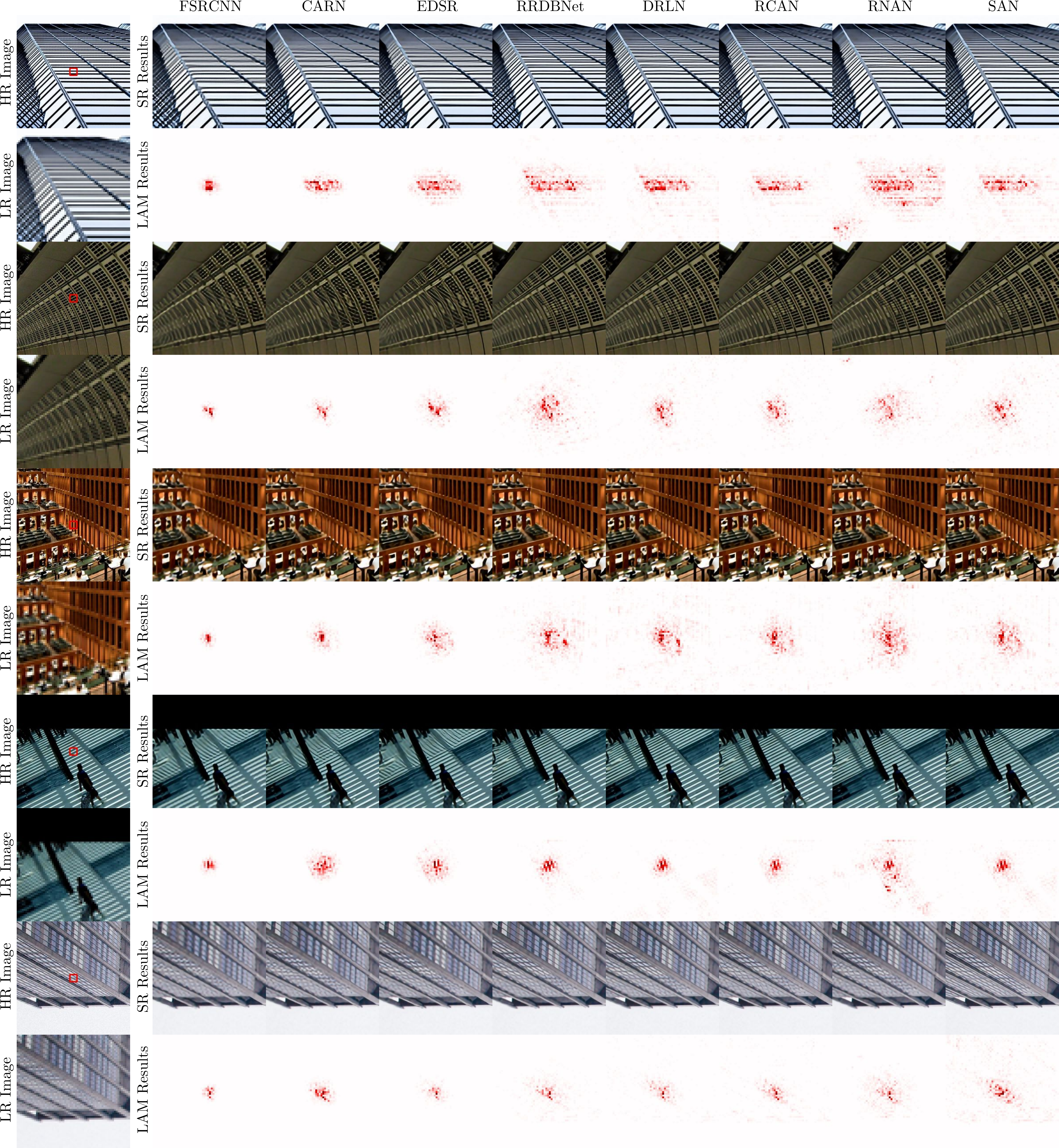}
    \caption{Comparison of the SR results and LAM attribution results of different SR networks. The LAM results visualize the importance of different pixel \wrt the SR results.}
    \label{fig:apd:lam_results_4}
\end{figure*}

In this section, we exhibit more results.
We first show more examples of the ``area of interest''.
In \figurename~\ref{fig:area_of_interest}, we have shown the five images with the smallest area of interest and also five images with the largest area of interest.
In \figurename~\ref{fig:apd:area_of_interest}, we show more images with their area of interest and the rank indices are also marked.
In \figurename~\ref{fig:apd:lam_results_1}, \figurename~\ref{fig:apd:lam_results_2}, \figurename~\ref{fig:apd:lam_results_3}, and \figurename~\ref{fig:apd:lam_results_4}, we show more LAM results.

\section{Change Log}
\noindent\textbf{v1}\quad initial pre-print release.

\noindent\textbf{v2}\quad We correct a typo that there should not be a $\frac{1}{m}$ at the end of Eq~\eqref{eq:PathIntegratedGradientApprox}. There is no such problem with the code. The conclusions of this paper are not affected.

\end{document}